\documentclass[citeauthoryear]{llncs}

\usepackage{graphicx}
\usepackage{amsmath,amssymb} 
\usepackage{color}
\usepackage[width=122mm,left=12mm,paperwidth=146mm,height=193mm,top=12mm,paperheight=217mm]{geometry}
\usepackage[pagebackref=true,breaklinks=true,letterpaper=true,colorlinks,bookmarks=false]{hyperref}

\usepackage{caption}
\usepackage{subcaption}
\usepackage{color,soul}
\usepackage[dvipsnames]{xcolor}
\usepackage{animate}
\usepackage{multirow}
\usepackage{enumitem}

\begin{document}
\pagestyle{headings}
\mainmatter

\title{Is Image Memorability Prediction Solved?} 
\author{Shay Perera, Ayellet Tal, Lihi Zelnik-Manor}
\institute{The Technion - Isreal}

\maketitle



\newcommand*{\ShowNotes}{}
\ifdefined\ShowNotes
  \newcommand{\colornote}[3]{{\color{#1}\bf{#2: #3}\normalfont}}
\else
  \newcommand{\colornote}[3]{}
\fi
\newcommand {\lihi}[1]{\colornote{magenta}{Lihi}{#1}}
\newcommand {\shay}[1]{\colornote{blue}{Shay}{#1}}
\newcommand {\ayellet}[1]{\colornote{red}{Ayellet}{#1}}
\newcommand{\etal}{{\em et al.}}
\newcommand{\setup}[3]{{{\color{teal}{#1}}-{\color{violet}{#2}}+{\color{olive}{#3}}\normalfont}}
\newcommand{\cnn}[1]{{{\color{teal}{#1}}\normalfont}}
\newcommand{\traindata}[1]{{{\color{violet}{#1}}\normalfont}}
\newcommand{\finetune}[1]{{{\color{olive}{#1}}\normalfont}}

\setlength{\tabcolsep}{2pt} 

\begin{abstract}
This paper deals with the prediction of the memorability of a given image.
We start by proposing an algorithm that reaches human-level performance on the LaMem dataset---the only large scale benchmark for memorability prediction. 
The suggested algorithm is based on three observations we make regarding convolutional neural networks (CNNs) that affect memorability prediction. 
Having reached human-level performance we were humbled, and asked ourselves whether indeed we have resolved memorability prediction---and answered this question in the negative. 
We studied a few factors and made some recommendations that should be taken into account when designing the next benchmark.
\end{abstract}

\section{Introduction}
\label{sec:Intro}
Our lives wouldn't be the same if we were unable to store visual memories. 
The vast majority of the population relies heavily on visuals to identify people, places and objects. Interestingly, despite different personal experiences, people tend to remember and forget the same pictures~\cite{bainbridge2013intrinsic,isola2014makes}.
This paper deals with the ability of algorithms to assess the memorability of a given image. 
Good memorability prediction could be useful for many applications, such as improving education material, storing for us things we tend to forget, producing unforgettable ads, or even presenting images in a way that is easier to consume.

\begin{figure*}[h]
\centering
\includegraphics[width=\textwidth]{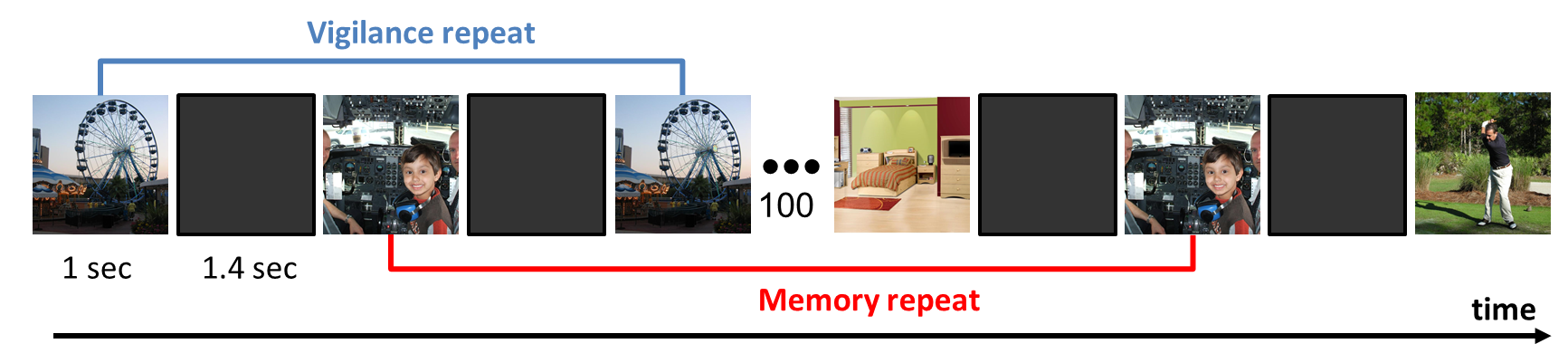}
\caption{{\bf Visual Memorability Game.} Participants watch for repeats in a long stream of images.}
\label{fig:game}
\end{figure*}

Image memorability is commonly measured as the probability that an observer will detect a repetition
of a photograph a few minutes after exposition, when presented
amidst a stream of images~\cite{bainbridge2013intrinsic,isola2014makes,isola2011makes,khosla2015understanding,borkin2013makes,dubey2015makes}, as illustrated in Figure~\ref{fig:game}.
According to cognitive psychological studies, this measurement determines which images left a trace in our long-term memory~\cite{isola2011makes,brady2008visual,konkle2010scene,standing1973learning}.

Several methods for memorability prediction have been proposed over the years~\cite{isola2011makes,isola2011understanding,mancas2013memorability,kim2013relative}.
A key observation is that both the type of scene and the type of objects in the image are highly related to its memorability~\cite{isola2014makes,khosla2015understanding,dubey2015makes}.
Based on this observation, Khosla \etal~\cite{khosla2015understanding} collected  a large-scale dataset, called {\em LaMem}, and proposed {\em MemNet} 
for memorability prediction.

In this paper we describe a system that achieves what seem to be astonishing results---reaching the limit of human performance on {\em LaMem}.  
Does this mean that image memorability prediction is a solved problem? 
To answer this question, we look deeper into the factors that impact human performance.
We discuss some factors that have been overlooked when building the existing datasets for image memorability. 
Our observations may lead to further studies not only in the design of meaningful datasets and effective algorithms for memorability prediction, but also in finding additional factors influencing memorability.

In the first part of the paper (Section~\ref{sec:memPrediction}), we propose a framework, {\em MemBoost}, for predicting image memorability, which achieves state-of-the-art results on all existing datasets.
%
This is done by delving into the relation between networks for image classification and memorability prediction. 
Our study gives rise to three main insights on which we base the design of {\em MemBoost}:
(i) As object classification CNNs improve, so does memorability prediction.
(ii) Scene classification plays a bigger role in memorability prediction than object classification.
This resolves conflicting opinions on the matter.
(iii) It suffices to train a regression layer on top of a CNN, which is designed and trained for object \& scene recognition, to achieve on par results with those attained by re-training the entire CNN for memorability prediction.
This insight contradicts previous observations.


Since our prediction results are surprising, in the second part of the paper (Section~\ref{sec:validity}) we re-visit some aspects that influence human performance in the memory game.
Via empirical analysis we show that changing some of the design decisions in the data collection could lead to data that better represents human memorability. 
The main conclusion from this study is that reaching human performance on LaMem does \emph{not} mean that memorability prediction has been solved. 
We further provide guidelines for building future datasets. 

In summary, this paper makes three major contributions.
First, it presents insights that should be the basis for memorability prediction algorithms.
A couple of these insights have already been demonstrated for other tasks in computer vision.
Second, the paper suggests a new framework that achieves state-of-the-art results, reaching the limit of human performance on LaMem.
Third, we put in question the portrayal of human memorability by the current datasets and give some recommendations towards the creation of the next large dataset.

\section{Previous Work}
\label{sec:related-work}
We start by describing how the ground-truth data has been collected.
We then review image attributes that have been studied with regard to memorability.

\vspace{-0.2in}
\subsubsection{The memory game.}
\label{subsec:game}
Image memorability is commonly measured using a memory-game approach, which was originally proposed by Isola \etal~\cite{isola2011makes}.
Briefly, the participants view a sequence of images, each of which is displayed for a predefined period of time, with some gap in between image presentations, as illustrated in Figure~\ref{fig:game}. 
The task of the participants is to press a button whenever they see an identical repeat of an image at any time in the
sequence~\cite{brady2008visual,konkle2010scene}. 
The participants receive feedback (correct or incorrect) whenever they press a key.

Unbeknown to the participants, the sequence of images is composed of “targets” and “fillers”, both are randomly sampled from the dataset. 
The role of the fillers is two-fold. 
First, they provide spacing between the first and the second repetition of a target. 
Second, responses on repeated fillers constitute a “vigilance task” that allowed us to continuously
check that the participants are attentive to the task [11, 12]. 
Each target is repeated exactly once and each filler is presented at most once (vigilance-task fillers are sequenced to repeat exactly once).

The memorability score assigned to each target image is the percentage of correct detections by the participants.
Throughout this paper, we refer to the memorability scores collected through the memory game as ground truth.
It is believed that by randomizing the sequence each participant sees, the measurements depend only on factors that are intrinsic to the image, independent of extrinsic variables, such as display order, time delay, and local visual context. 


\vspace{-0.2in}
\subsubsection{Observations.}
Since our study deals with the relation between scenes, objects, and memorability, we next provide a brief review of previous observations regarding which image attributes affect (or do not affect) image memorability.

\noindent 
1. \textit{Object category and scene category attributes~\cite{isola2014makes,isola2011makes,dubey2015makes,isola2011understanding,kim2013relative,khosla2012memorability,bylinskii2015intrinsic}.}
A bunch of studies on this topic concluded that some object categories, such as people, vehicles, and animals, and some scene categories, such as indoor scenes, are more memorable than others.

\noindent 
2. \textit{Semantic attributes.}
Scene semantics go beyond just content and scene category~\cite{isola2014makes,isola2011makes,isola2011understanding}. 
Features such as spatial layout or actions are highly correlated with memorability and are an efficient way of characterizing memorability.

\noindent 
3.  \textit{Saliency.}
While Khosla \etal~\cite{khosla2012memorability} and Celikkale \etal~\cite{celikkale2015predicting} show a reasonable  correlation between memorability and attention, Mancas~\etal~\cite{mancas2013memorability} show almost no correlation between the two.
Furthermore, Khosla \etal~\cite{khosla2015understanding} found a reasonable correlation between fixation duration to memorability.
When considering object memorability, rather than image memorability, Dubey \etal~\cite{dubey2015makes} found high correlation with  the number of unique fixation points within the object.

\noindent 
4. \textit{Object statistics~\cite{isola2014makes,isola2011makes,isola2011understanding}.} 
The number of objects and the object area have low correlation with memorability and are ineffective at predicting memorability.

\noindent 
5. \textit{Aesthetics, interestingness, emotion and popularity.}
Both image aesthetics and interestingness show no correlation to memorability~\cite{isola2014makes,khosla2015understanding}, while they are correlated with each other. 
Popularity is correlated to memorability only for the most
memorable images, but not for others~\cite{khosla2015understanding}.
%
Negative emotions, such as anger and fear, tend to be more memorable than those portraying positive
ones \cite{khosla2015understanding}.

\noindent 
6. \textit{Colors~\cite{isola2014makes,isola2011makes,isola2011understanding,konkle2010conceptual,george10real}.} 
Colors have only weak correlation with memorability. 

\noindent 
7.  \textit{What people think is memorable~\cite{isola2014makes}.}
Asking people to guess which images are the most memorable in a collection reveals low correlation to the actually memorable ones.
Participants have wrong intuition, erroneously assuming that beautiful and interesting images will produce a lasting memory.


\section{MemBoost: A System for Predicting Image Memorability}
\label{sec:memPrediction}


Motivated by the strong evidence for the correlation between scenes, objects and image memorability, we explored the utility of this correlation for training CNNs.
We start our study with short descriptions of three insights we make on memorability prediction with CNNs.
Then, we propose the \emph{MemBoost} algorithm, which is based on these insights, and provides state-of-the-art results.

Our experimental setup follows that of~\cite{khosla2015understanding} for a variety of datasets and for a variety of networks.
Briefly, we randomly split the images in each memorability dataset into a train set and a test set. 
We performed experiments on three datasets: LaMem~\cite{khosla2015understanding}, SUN-Mem~\cite{isola2011makes}, and Figrim~\cite{bylinskii2015intrinsic}.
We note that LaMem is a relatively large dataset, consisting of $58,741$ images, whereas the other datasets consist of only $2,222$ and $1,754$ images, respectively; see the appendix for details on these dataset (Table~\ref{tbl:datasets}).
For SUN-Mem and Figrim, splitting was repeated for $25$ times, while for LaMem, we used $5$ splits due to its size.

In order to measure the prediction performance, we follow the common practice in the evaluation of memorability prediction.
That is to say, rather than comparing memorability scores directly, we compare ranks as follows.
The images in the test set are ranked both according to their ground-truth memorability scores and according to the algorithm predictions.
The Spearman's rank correlation (\(\rho\)) is computed between the two rankings.

\subsection{Insights}
\label{subsec:insights}

In this section we suggest three insights and explore their validity, one by one, via thorough experiments.

\vspace{-0.1in}
\subsubsection{1. Use a strong base CNN.}
It has recently become common knowledge that the stronger your backbone CNN is, the better results you'd get, even if the base CNN was not trained for your specific task.
In accordance with this, we compare in Table~\ref{tbl:classNets} four architectures: ResNet152~\cite{he2016deep}, VGG16~\cite{simonyan2014very}, GoogLeNet~\cite{szegedy2015going} and AlexNet~\cite{krizhevsky2012imagenet}.
All of them were trained on ImageNet~\cite{deng2009imagenet} and XGBoost was used for the regression layer.
The table shows that indeed, as classification networks improve, so do the corresponding memorability prediction networks.
These results are consistent across all three datasets.
This implies that it might be unnecessary to develop special architectures for memorability prediction.
Instead, it suffices to update the relevant layers of the best-performing classification network.

\begin{table*}[!ht]
\begin{center}
\small{
\begin{tabular}{| l | c | c | c | c |c|}
\hline
\textbf{Setup} & \multicolumn{3}{ c |}{\textbf{Memorability Prediction}}  & \multicolumn{2}{ c |}{\textbf{Classification}} \\ 
\cline{2-6}
\textbf{Vary Network} & \textbf{LaMem} & \textbf{SUN-Mem} & \textbf{Figrim} &  \textbf{SUN }  &  \textbf{Places}\\
\hline
\setup{ResNet152}{ImageNet}{XGBoost}  & \textbf{0.64} & 0.64 & \textbf{0.56} & - & 54.74\\ \hline
\setup{VGG16}{ImageNet}{XGBoost} & \textbf{0.64} & \textbf{0.65} & \textbf{0.56} & 48.29 & 55.24
\\ \hline
\setup{GoogLeNet}{ImageNet}{XGBoost} & 0.62   & 0.63 & 0.55 & 43.88 & 53.63\\ \hline
\setup{AlexNet}{ImageNet}{XGBoost} & 0.61 & 0.6 & 0.51 & 42.61 & 53.17\\ \hline
\end{tabular}
}
\end{center}
\centering
\caption{{\bf Memorability prediction improves with classification.} 
The table shows results of two tasks: image memorability prediction and image classification. 
For classification, we show results on two datasets: SUN~\cite{xiao2010sun} and Places~\cite{zhou2014learning}.
The accuracy is computed for four architectures, all trained on ImageNet.
Consistently, as classification accuracy improves, so does memorability prediction.
We adopt the following naming and color-code convention for the setups:
\setup{Network}{Training dataset}{Regression type}.
\vspace{-0.3in}
}
\label{tbl:classNets}
\end{table*}

\vspace{-0.15in}
\subsubsection{2. Training on scene classification is more important than training on object classification.}
Our second insight regards the essence of memorability:
What makes an image memorable, the objects in it or the scene it describes?
Answering this question is not only interesting theoretically, but it also has very practical implications, since it will enable a better selection of the training dataset for memorability prediction.   

Previous works have found that scene category and object presence are, together, highly correlated with the memorability of an image 
(Spearman rank correlation \(\rho=0.43\))~\cite{isola2011makes,isola2011understanding,khosla2012memorability}.
In~\cite{isola2014makes} it is claimed that this correlation is mostly due to the scene category itself, which appears to summarize much of what makes an image memorable.
However, this observation has not been used for memorability prediction.


We empirically verify this claim and show, in Table~\ref{tbl:objectScene}, that CNNs that are trained on datasets of scenes (Places205~\cite{zhou2014learning}) outperform CNNs that are trained on datasets of objects (ImageNet~\cite{deng2009imagenet}).
However, the accuracy is not as good as that obtained when training on both objects and scenes (Hybrid1205~\cite{zhou2014learning} or {Hybrid1365}~\cite{zhou2017places}).
This behavior is persistent across datasets and networks (results are shown for both AlexNet and ResNet152).


\begin{table}[htb]
\begin{center}
\small{
\begin{tabular}{| l | c | c | c |}
\hline
\textbf{Setup} & \multicolumn{3}{ c |}{\textbf{Memorability Prediction}} 
\\ 
\cline{2-4}
\textbf{Vary training data} & \textbf{LaMem} & \textbf{SUN-Mem} & \textbf{Figrim}
\\
\hline
\setup{AlexNet}{ImageNet}{XGBoost} & 0.61 & 0.6 & 0.51 
\\
\setup{AlexNet}{Places205}{XGBoost} & 0.61 & 0.64 & 0.55 
\\ 
\setup{AlexNet}{Hybrid1205}{XGBoost}  & \textbf{0.64} & \textbf{0.65} & \textbf{0.57}  \\\hline
\setup{ResNet152}{ImageNet}{XGBoost} & 0.64 & 0.64 & 0.56
\\
\setup{ResNet152}{Places365}{XGBoost} & 0.65 & \textbf{0.66} & 0.56 
\\
\setup{ResNet152}{Hybrid1365}{XGBoost} & \textbf{0.67 } & \textbf{0.66} & \textbf{0.57} \\\hline
\end{tabular}
}
\end{center}
\centering\caption{{\bf Scenes are more important than objects.}
Memorability was predicted using \cnn{AlexNet} and \cnn{ResNet152}, trained on objects \traindata{(ImageNet)}, on scenes \traindata{(Places205 \& Places365)}, or on their combination \traindata{(Hybrid1205 \& Hybrid1365)}. 
While scenes are more important than objects, their combination slightly improves the prediction. (The training datasets' details are given in the appendix).
}
\label{tbl:objectScene}
\end{table}

We note that these datasets do not provide statistics regarding the balance between scenes and objects.
Section~\ref{sec:validity} discusses the validity of the training datasets.
One such consideration is already clear---the right balance between objects and scenes  should be sought after.

\vspace{-0.2in}
\subsubsection{ 3. Re-training may be unnecessary.}
This insight regards the training of classification networks with memorability data.
Is it really necessary to fine-tune the entire network or is it sufficient to train just the last regression layer? 
Answering this question in the affirmative means that we can achieve good results even when we have neither a lot of memorability data nor much computational resources for training.
This is important since such data is not widely available and is difficult to collect, whereas classification data is more widespread.

Khosla \etal~\cite{khosla2015understanding} compare two approaches for re-training a CNN for memorability prediction.
The first approach re-trains only the last regression layer using Support Vector Regression (SVR).
In the second approach, called {\it MemNet}, the entire network is fine-tuned with memorability data. 
They achieve better results with MemNet and conclude that fine-tuning the entire CNN is essential.

We reach the opposite conclusion. 
We show that modifying only the regression layer can provide comparable memorability prediction to re-training the entire network. 
In particular, we took the same network setup as~\cite{khosla2015understanding}, using AlexNet trained on Hybrid1205.
We eliminated the classification layer {\em (top softmax layer)} and considered the previous layer as features.
We then replaced the classification layer by training a regressor model, which is based on boosted trees (using XGBoost library~\cite{chen2016xgboost}) and maps the features to memorability scores.

As can be seen in Table~\ref{tbl:tuning}, fine-tuning only the regression layer with XGBoost is a good idea.
In particular, on the large-scale LaMem, training just the regression layer yields the same accuracy as MemNet ($0.64$ in both cases). 
On the smaller dataset SUN-Mem, training just the regression layer even gives better results than re-training the entire network ($0.65$ in comparison to $0.53$). 

\begin{table}[htb]
\begin{center}
\small
\begin{tabular}{| l | l | c | c | c | c |}
\hline
& \textbf{Training approach} & \textbf{LaMem} & \textbf{SUN-Mem} & \textbf{Figrim} \\
\hline\hline
& Human consistency & 0.68 & 0.75 & 0.74 \\ 
\hline\hline
\multirow{1}{*}{\cite{khosla2015understanding}}  & \setup{AlexNet}{Hybrid1205}{SVR} & 0.61 & 0.63 & - \\ 
\cline{2-5}
[MemNet] & \setup{AlexNet}{Hybrid1205}{Fine-tune}  & \textbf{0.64} & 0.53 & - \\ 
\hline\hline
Ours &\setup{AlexNet}{Hybrid1205}{XGBoost}  & \textbf{0.64} & \textbf{0.65} & \textbf{0.57} \\ \cline{2-5}
\hline
\end{tabular}
\end{center}
\centering\caption{ {\bf Fine-tuning may be unnecessary.} Modifying only the regression layer provides comparable memorability prediction to re-training the entire network. 
The setup used by~\cite{khosla2015understanding} is utilized.}
\vspace*{-0.3cm}
\label{tbl:tuning}
\end{table}

\subsection{The MemBoost algorithm}
\label{subsec:SOTA}

\begin{figure}[htb]
\centering
\includegraphics[width=0.9\textwidth]{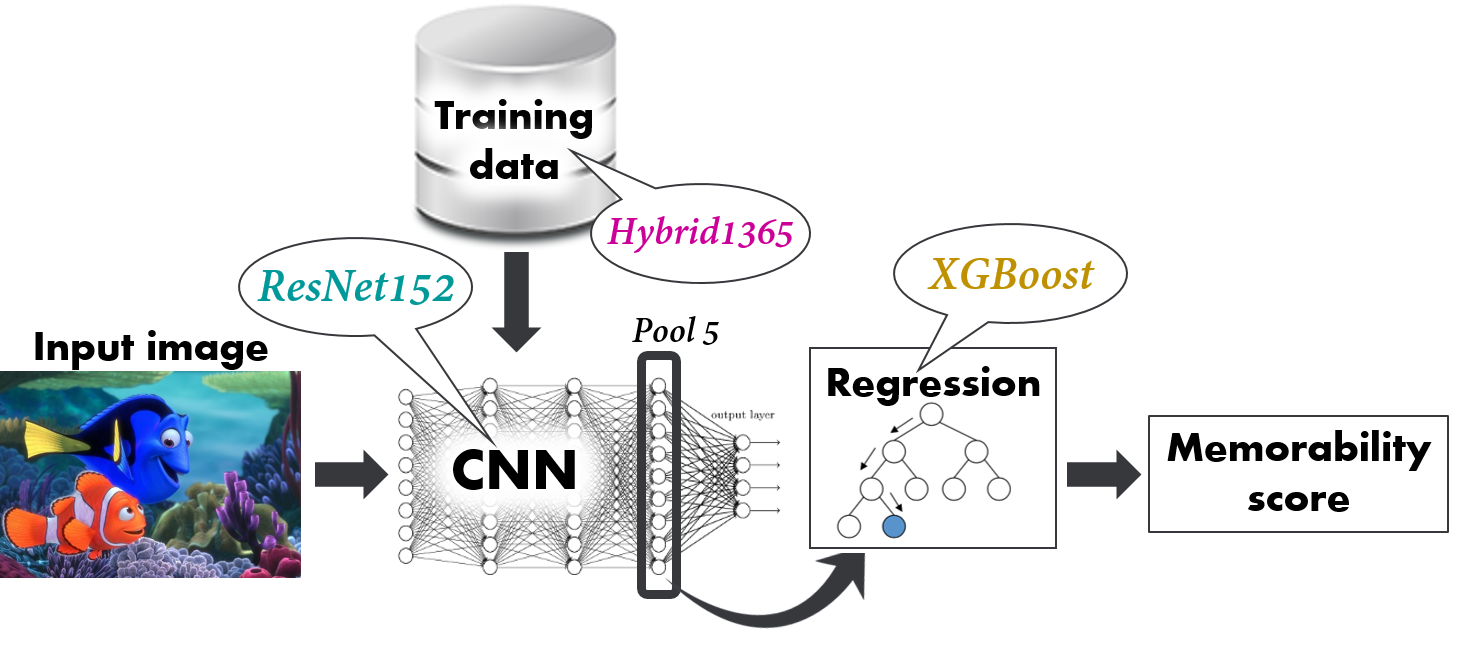}
\caption{{\bf The MemBoost pipeline.} 
Deep features are extracted from {\em pool5} layer of ResNet152, trained on Hybrid1365 dataset. 
A boosted-trees regression model is trained using these features to predict image memorability.
}
\vspace*{-0.3cm}
\label{fig:algorithm}
\end{figure}

Our next step is to utilize the observations from Section~\ref{subsec:insights} to design a novel algorithm, called {\em MemBoost}.
As suggested by insight (1), we select ResNet152 as our base network.
We follow insight (2) and use a version named ResNet152-Hybrid1365 that was trained both on an object dataset (ImageNet~\cite{russakovsky2015imagenet}) and on a scene dataset (Places365~\cite{zhou2017places}).
Last, in sync with insight (3), we modify only the regression layer, using XGBoost~\cite{chen2016xgboost}, to map deep features to memorability scores. 
%
Figure~\ref{fig:algorithm} illustrates the pipeline of our approach for acquiring state-of-the-art memorability prediction.

Table~\ref{tbl:SOTA} summarizes our results.
The top row of the table presents the memorability consistency across different groups of human observers. 
This serves as an upper bound.
The next two rows show the previous best results, obtained by the two approaches of Khosla \etal~\cite{khosla2015understanding}. 
The bottommost row of the table shows the results of our MemBoost, which closes the gap with human prediction results on the LaMem dataset.
MemBoost provides a significant improvement over MemNet on both LaMem and SUN-Mem (\cite{khosla2015understanding} did not test on Figrim and we failed to reproduce their training).

\begin{table}[htp]
\begin{center}
\small
\begin{tabular}{| l l | l | c | c | c | c |}
\hline
 && \textbf{Approach} & \textbf{LaMem} & \textbf{SUN-Mem} & \textbf{Figrim} \\
\hline\hline
& & Human consistency & 0.68 & 0.75 & 0.74 \\ 
\hline\hline
\multirow{2}{*}{\cite{khosla2015understanding}} & & \setup{AlexNet}{Hybrid1205}{SVR} & 0.61 & 0.63 & - \\ 
\cline{2-5}
& [MemNet] & \setup{AlexNet}{Hybrid1205}{Fine-tune}  & 0.64 & 0.53 & - \\ 
\hline\hline
 Our & [MemBoost] & \setup{ResNet152}{Hybrid1365}{XGBoost} & \textbf{0.67} & \textbf{0.66} & \textbf{0.57} \\ 
\hline
\end{tabular}
\end{center}
\centering\caption{ {\bf Memorability prediction results.} 
The table compares our MemBoost algorithm results with those of~\cite{khosla2015understanding}. 
It shows that our insights lead to state-of-the-art memorability prediction on all three datasets.}
\vspace*{-0.3cm}
\label{tbl:SOTA}
\end{table}

\begin{figure*}[htp]
\centering
\begin{tabular}{cc}
\includegraphics[height=4.8cm]{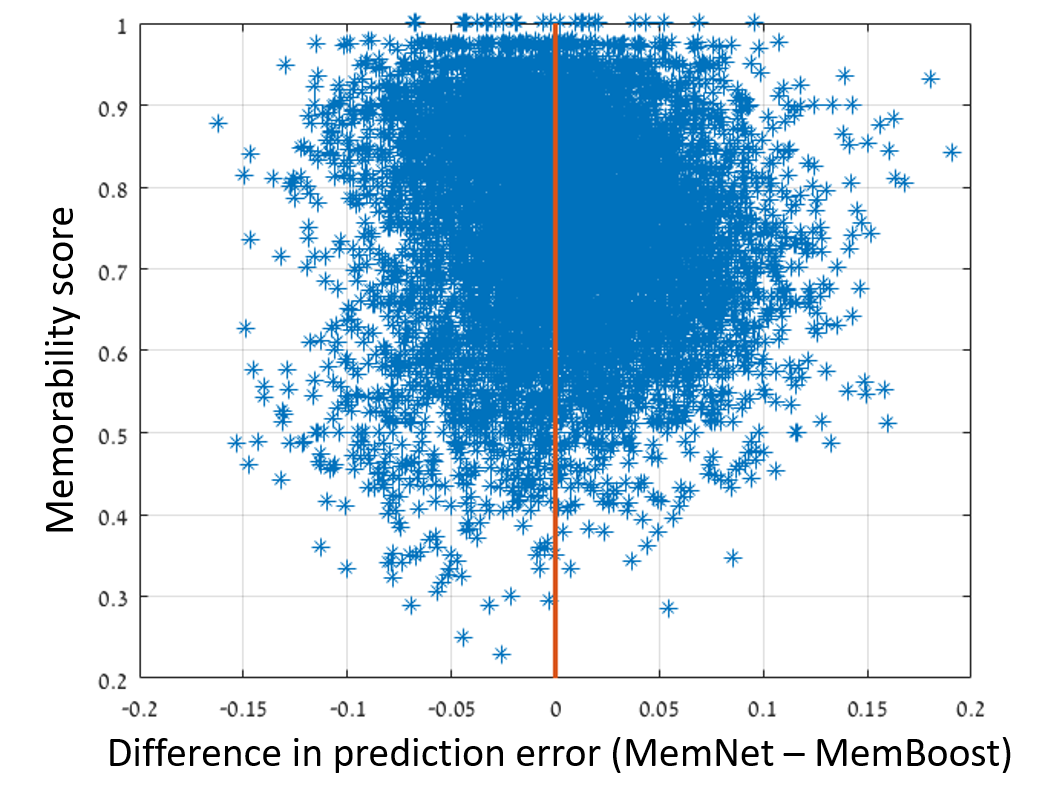}&
\includegraphics[height=4.8cm]{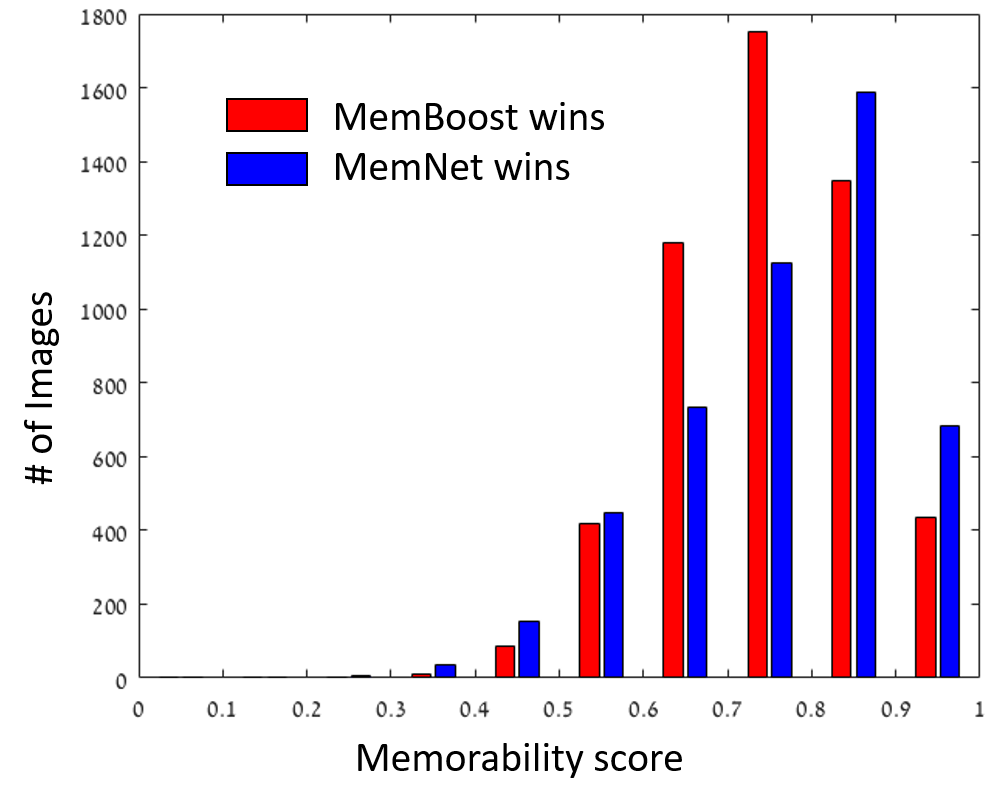}\\
(a) & (b)
\end{tabular}
\caption{{\bf MemBoost vs. MemNet}.
(a) Each point represents an image. The Y-axis shows memorability scores and the X-axis shows the difference in prediction error between MemBoost and MemNet.
MemBoost is better when this difference is positive (right of the red line that represents equal-error).
The distribution is not symmetric and shows that in the [0.6,0.8] range, i.e., when the images are not highly memorable or highly forgettable, MemBoost is more accurate.
(b)~Histograms of the number of images where MemBoost is more accurate vs. the number of images where MemNet is more accurate, as a function of the memorability score.
It is evident that MemBoost is preferable in the mid-memorability range, while MemNet is more accurate for highly memorable images.
Overall, MemBoost's results outperform those of MemNet.
}
\label{fig:predError}
\end{figure*}


%
%

\vspace{-0.1in}
\subsubsection{Result analysis.}
Figure~\ref{fig:predError} sheds light on where our predictions are more accurate than those of MemNet.
Every point in the graph in Figure~\ref{fig:predError}(a) corresponds to an image.
The Y-axis shows the memorability score of the image.
The X-axis shows the difference between MemBoost prediction error and MemNet's.
When MemBoost outperforms MemNet, this difference is positive and vice versa.
A symmetric distribution of points around the equal-error line (X=0) would mean both methods have similar distributions of errors.
As can be seen, on LaMem the distribution looks more like a Pac-Man, with its mouth at medium memorability scores (0.6-0.8).
In Figure~\ref{fig:predError}(b) a different view of the same behavior is given. 
This stands to show that when the images are not highly memorable or highly forgettable, our algorithm wins.
We believe that this is so since such images are more challenging as the scenes/objects are less recognizable.
A more powerful prediction algorithm is hence essential in these cases.

\begin{figure*}[tb]
\centering
\begin{tabular}{rccc}
 \includegraphics[height=3.1cm]{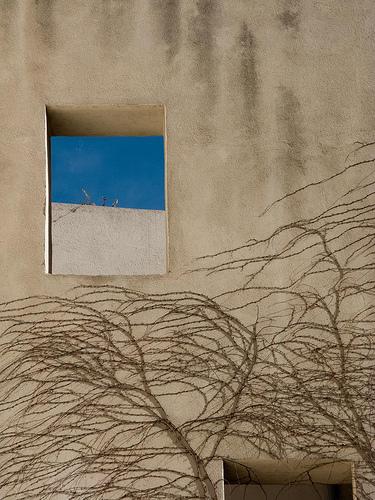} &
 \includegraphics[height=3.1cm]{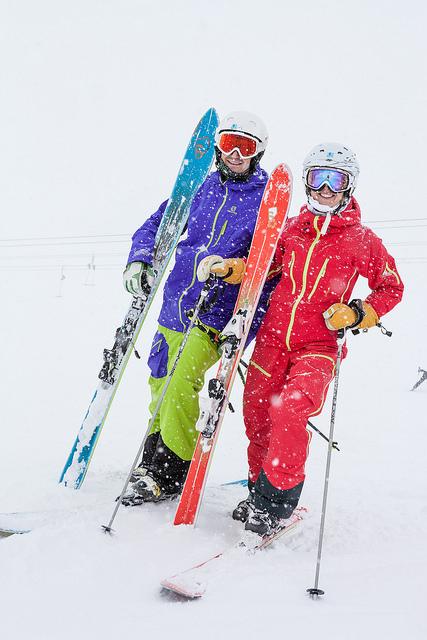} &
 \includegraphics[height=3.1cm,]{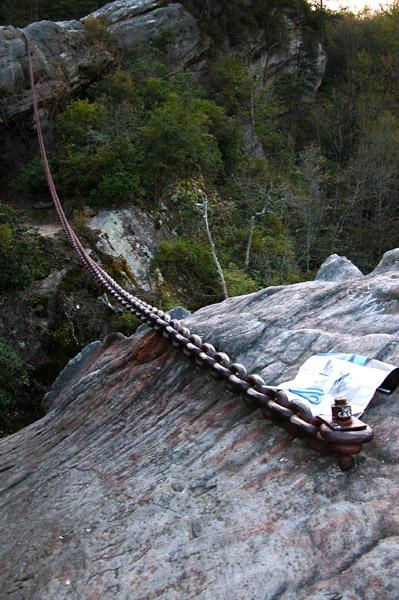} &
 \includegraphics[height=3.1cm]{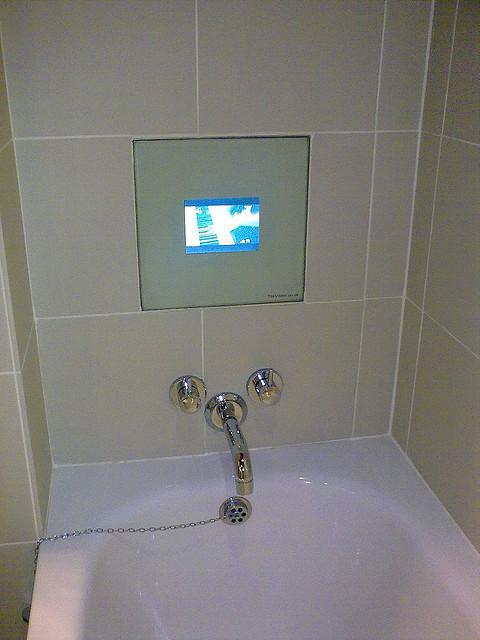} \\
{\small GT: (0.63;1583)} &
{\small (0.70;3150) } &
{\small (0.79;5871) } &
{\small (0.68;2571) } \\
{\small MemNet: (0.84;7623) } &
{\small(0.86;8587)} &
{\small (0.65;1393) } &
{\small (0.86;8301) } \\
{\small  MemBoost: (0.72;2967)} &
{\small (0.75;4295) } &
{\small (0.78;5899)} &
{\small (0.74;4082)} \\\\
 \includegraphics[height=2.0cm]{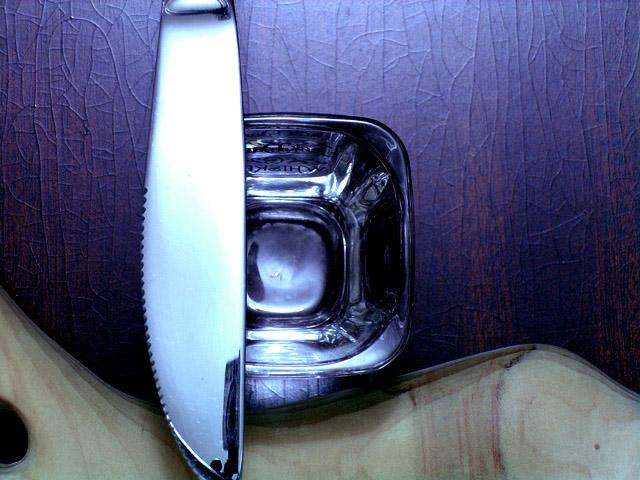} &
 \includegraphics[height=2.0cm]{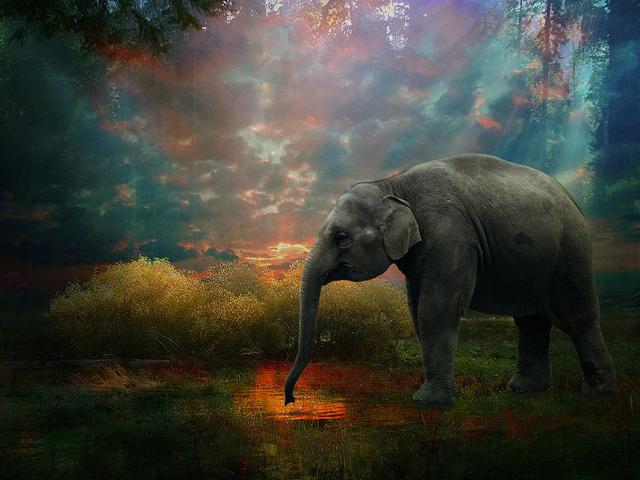} &
 \includegraphics[height=2.0cm]{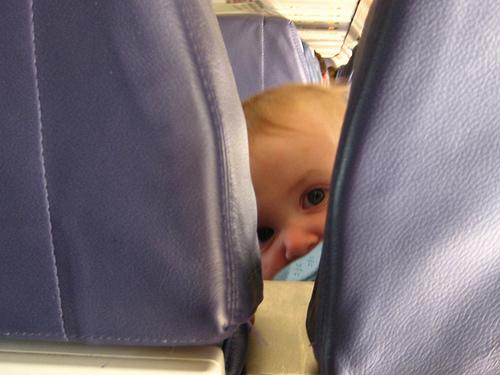} &
 \includegraphics[height=2.0cm]{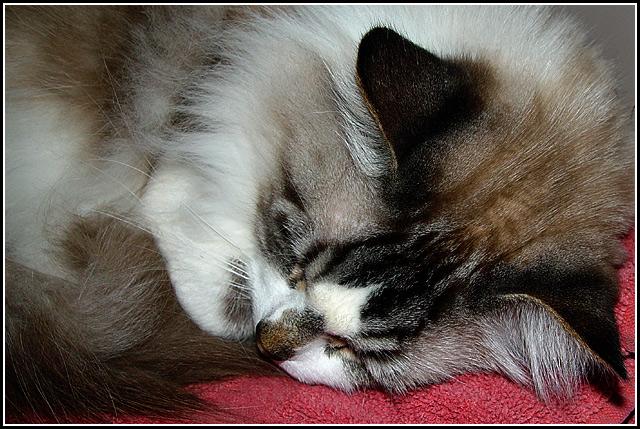} \\
{\small GT: (0.65;2082) }&
{\small (0.80;6139) } &
{\small (0.74;4332) }&
{\small (0.66;2240) }\\
{\small  MemNet: (0.88;9068) }&
{\small (0.63;947) } &
{\small (0.89;9330) }&
{\small (0.82;6901) }\\
{\small MemBoost:   (0.75;4515)}&
{\small (0.77;5195)} &
{\small (0.77;5174)}&
{\small (0.70;2511)}\\

\end{tabular}
\caption{{\bf Qualitative results of images with medium memorability scores.}
The memorability scores of these images are probably due to memorable objects and forgettable scenes (skiers, cats, baby, chain) or forgettable objects and memorable scenes (window, elephant, bathroom).
Our algorithm, which combines objects and scenes, manages to do better than its competitors.
For each image we present two values: its memorability score and its rank within the 10,000 images of Test\_1 set of LaMem. 
\vspace{-0.2in}
} 
\label{fig:predErrorImages1}
\end{figure*}

\begin{figure*}[htb]
\centering
\begin{tabular}{rccc}
 \includegraphics[height=2cm]{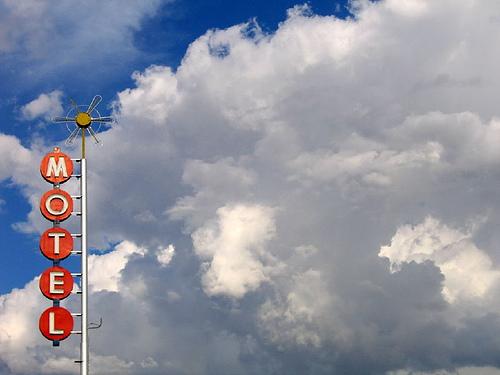} &
 \includegraphics[height=2cm]{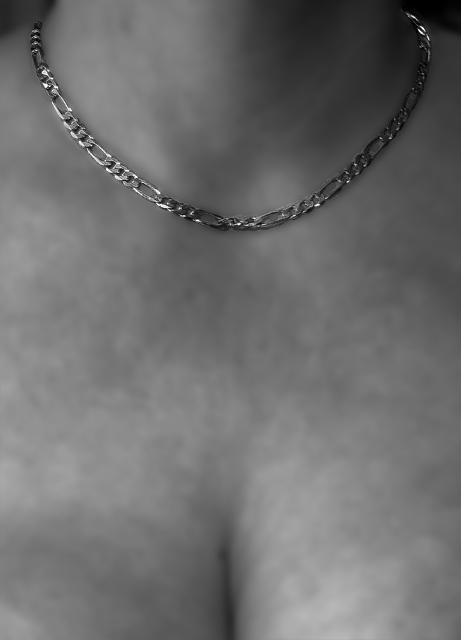} &
 \includegraphics[height=2cm]{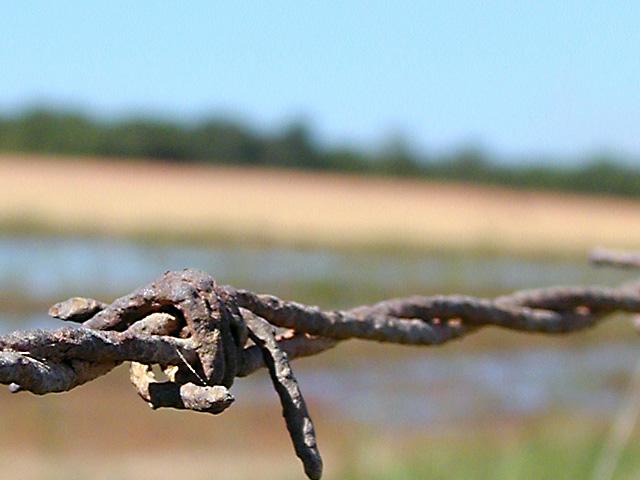} &
 \includegraphics[height=2cm]{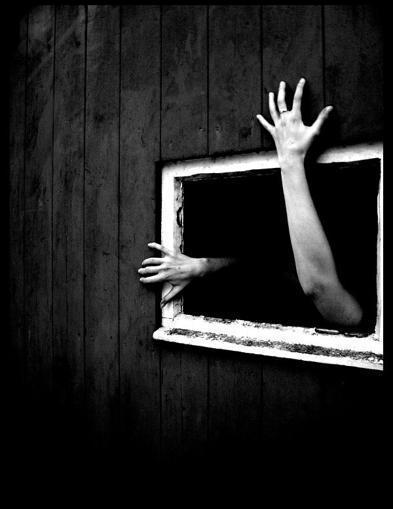} \\
{\small GT: (0.93;9511)} &
{\small (0.93;9547) }&
{\small (0.90;8884) } &
{\small (0.93;9405) }\\
{\small MemNet: (0.65;1357) } &
{\small (0.72;2979) }&
{\small (0.68;1862) } &
{\small (0.74;3689) }\\
{\small MemBoost: (0.83;8118)} &
{\small  (0.86;9070)}&
{\small (0.82;7648)} &
{\small (0.86;9105)}\\ \\
 \includegraphics[height=3.1cm]{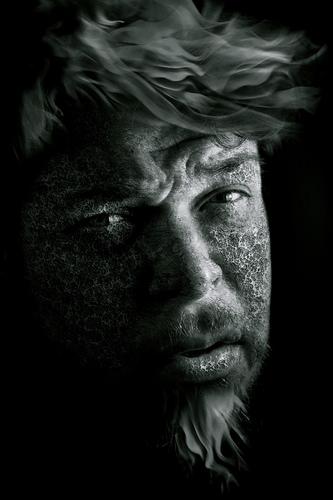} &
 \includegraphics[height=3.1cm]{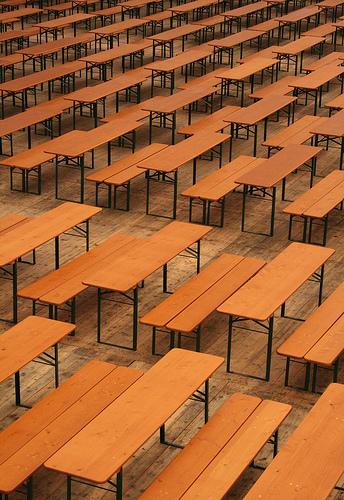} &
 \includegraphics[height=3.1cm]{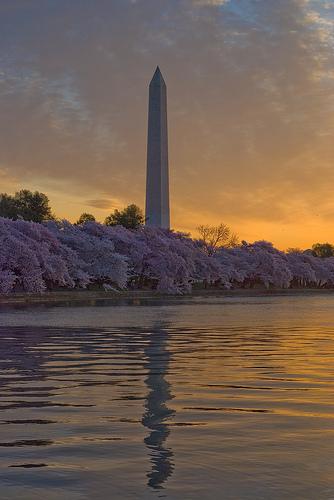} &
 \includegraphics[height=3.1cm]{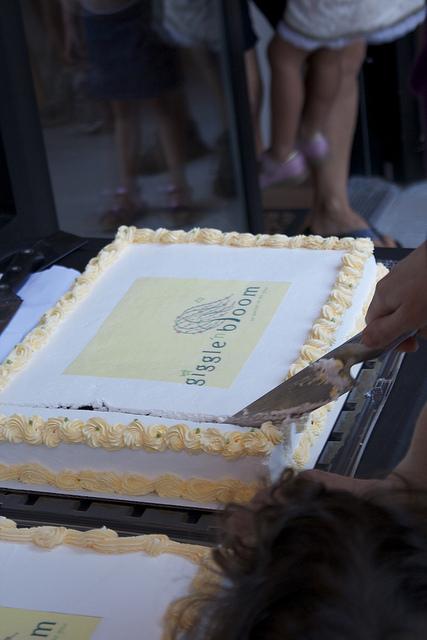} \\
{\small GT:  (0.89;8753) } &
{\small (0.92;9302) } &
{\small (0.84;7397) } &
{\small (0.85;7607) }\\
{\small MemNet: (0.73;3174)} &
{\small (0.76;4110) } &
{\small (0.63;948) } &
{\small (0.64;1196) }\\
{\small MemBoost: (0.84;8469)} &
{\small (0.87;9229)} &
{\small (0.79;6065)} &
{\small (0.79;6305)}\\

\end{tabular}
\caption{{\bf Qualitative results of images for which MemBoost outperforms MemNet by the largest margin.}
These images contain common objects in distinct scenes, for which MemNet failed to predict memorability accurately.
} 
\label{fig:predErrorImages2}
\end{figure*}

To complement our argument, we present in Figures~\ref{fig:predErrorImages1}~\&~\ref{fig:predErrorImages2} several example images and their corresponding scores.
 Figure~\ref{fig:predErrorImages1} shows images that have medium memorability scores, where MemBoost outperforms MemNet.
The common content of these images is having either memorable objects within forgettable scenes or forgettable objects within memorable scenes.

This observation is reinforced in Figure~\ref{fig:predErrorImages2}, which shows the images for which we got the largest gap in prediction accuracy between MemBoost and MemNet.
Interestingly, these images are all highly memorable, as evident from their memorability scores.
They all show common objects, but in unique scenes---text on a cake, hands coming out of a window, extreme repetition of objects, etc. 
Our algorithm, which combines scenes and objects, manages to predict that these are unique combinations, and hence memorable.

\section{On the Validity of Current Memorability Scores}
\label{sec:validity}

Recall that the upper bound on LaMem is the mean rank correlation between the memorability scores corresponding to different groups of people. 
As our solution reached this upper bound, we ask ourselves whether the memorability scores, obtained through the Memory Game described in Section~\ref{subsec:game}, form a sufficient representation. 
We explore three key questions and answer them via experiments:
\begin{enumerate}
\item Does the number of observers per image suffice as a representative sample?
\item How consistent are the scores across observers? Should the mean score be utilized by itself, or should the variance be considered as well?
\item Should the order in which the images are displayed be taken into account?
\end{enumerate}

\vspace{-0.1in}
\subsubsection{1. What is a sufficient sample set size?}

Recall that the memory game takes as scores the mean memorability over multiple participants. 
A key question then is `how many participants should attend the game for the mean scores to be meaningful?'

Isole \etal~\cite{isola2011makes} show that as the number of participants increases, the mean scores become more stable.
To prove stability they show that averaging memorability scores over groups of $40$ participants yields Spearman rank correlation of $\rho=0.75$ between different groups (on the SUN Memorability dataset).
Similarly, Khosla \etal~\cite{khosla2015understanding} measure rank correlation of $\rho=0.68$ on the Lamem dataset, and Bylinskii \etal~\cite{bylinskii2015intrinsic} measure rank correlation of $\rho=0.74$  on the Figrim dataset.

One problem with these results is that they ignore the standard deviation of the correlation when computed over different splits into groups. 
That is, how stable are the rank correlations between groups?
As it turns out,~\cite{bylinskii2015intrinsic} report quite a large variability across splits, i.e., $\sigma\!=\!0.2$ on Figrim~\cite{bylinskii2015intrinsic}.
This raises questions regarding the use of memorability scores for prediction, since falling within the variance should be considered as success.

We therefore aim at studying the best group size needed for consistent image memorability scores.
To do it, we repeated the Memory Game, as described in Section~\ref{subsec:game}, using target and filler images randomly selected from Figrim.
We evaluate human consistency across different group sizes as follows.
We measured memorability for 45 target images, each scored by 275 participants on average.
The participants were split into two equal-size groups and the mean score per image was computed for each group.
We then computed the Spearman’s rank correlation between the scores of the two groups.
This was done for 100 random splits into groups.
We then computed both the mean and the variance of the correlation scores over all splits.

\begin{figure}[htb]
\centering
    \includegraphics[width=0.6\textwidth]{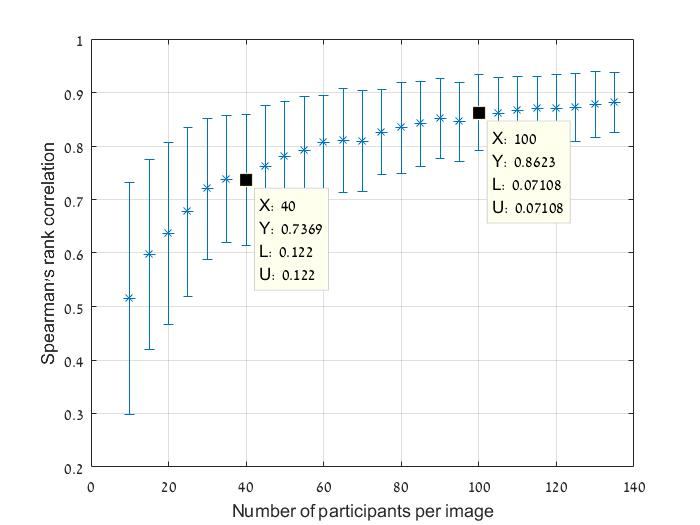}

\caption{{\bf Memorability consistency across groups of participans.} 
Using groups of $100$ participants is a good compromise between the accuracy of human consistency  and the complexity of collecting the data.}
\label{fig:consistency}
\end{figure}

Figure~\ref{fig:consistency} shows our results. 
For groups of $40$ participants, the consistency is $\rho=0.74$ with standard deviation of $\sigma=0.12$ (compared to $0.74$, $0.2$ respectively, reported in~\cite{bylinskii2015intrinsic}).
For groups of $100$ participants, the consistency significantly increases to $0.86$ ($\sigma=0.07$), while for groups of $135$ participants, it only slightly increases further to $0.88$ ($\sigma=0.05$).

We conclude that assigning memorability scores should be performed by averaging over groups bigger than $40$ participants.
Since collecting memorability measurements for large numbers of participants requires a great deal of work, we recommend using $100$ observers, which seems like a good compromise between the consistency and the complexity of collecting the data. 

\vspace{-0.2in}
\subsubsection{2. How consistent are the scores? Should the variance be considered?}

Having noted that the scores consistency varies, we further ask ourselves whether the mean scores, which are used by the existing datasets, have the same meaning for all images.
That is, we question whether representing an image by its mean memorability score suffices, as maybe the variance per image should also be considered.

\begin{figure}[htb]
\includegraphics[width=0.6\textwidth]{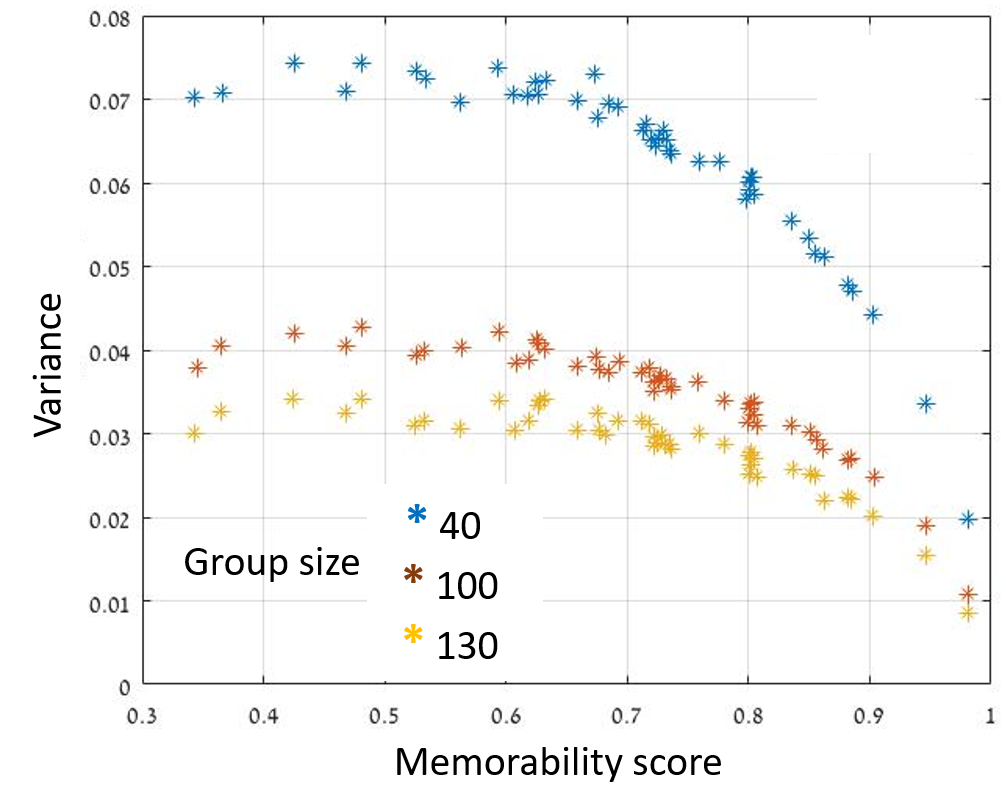}
\centering\caption{{\bf Memorability variance varies across images.} 
Some images are highly memorable by all, while for other images there is high variance. 
The variance decreases when computing scores over larger participant group size.}
\label{fig:variance}
\end{figure}

To answer this, we measured the variance of the memorability scores given by different groups of people, per image.
This was done for 45 images and for group sizes between $40$ and $130$.
%
Figure~\ref{fig:variance} shows our results.
Every point in the graph represents a different image.
The main conclusion from this graph is that the variance is not fixed.
Some images are highly memorable by most people, while others are memorable by some and not so much by others.
This suggests that one may want to represent image memorability using two numbers, the mean memorability score and the variance of the scores.

However, a second conclusion from Figure~\ref{fig:variance}, is that the larger the number of participants, the smaller the variance across groups. 
Therefore, if the number of observers is sufficiently large, it may suffice to maintain a single number---the memorability score---which is the common practice.
This supports our previous claim that a larger number of observers per image could lead to more stable scores, with higher consistency.

\vspace{-0.2in}
\subsubsection{3. Should the order in which the images are displayed be taken into account?}
\label{subsec:order}

In real life, images are always displayed in some context; for example, when looking at the newspaper or walking in a museum, the images intentionally appear in a certain order.
In the street, the order of images is not intentional, but it definitely  influences the scenes and objects we will remember or forget.
Image memorability, however, is commonly measured across random sequences of images in order to isolate extrinsic  effects such as the order of viewing.
Is that the right practice?
Should an image be assigned a single memorability score?

To assess the effect of image order on memorability, we designed a new version of the ``Memory Game'', in which the only difference from the original memory game is that rather than randomly creating image sequences, we used fixed sets of sequences.
For a fixed set of target and filler images (120 altogether, randomly chosen from Figrim), we created 5 orders.

\begin{figure*}[htb]
\centering
  \begin{subfigure}[b]{\textwidth}
  \includegraphics[width=\textwidth]{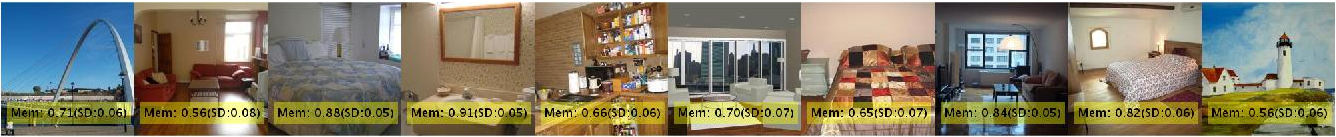}
  \end{subfigure}
  \begin{subfigure}[b]{\textwidth}
    \includegraphics[width=\textwidth]{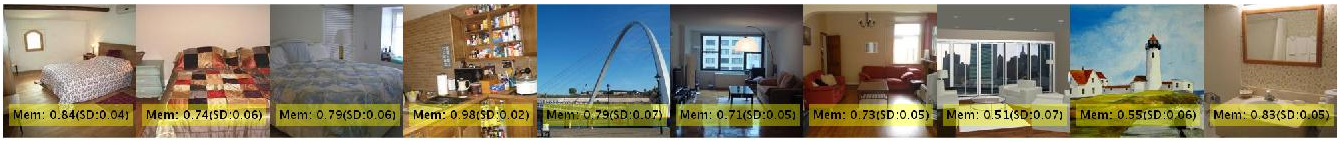}
  \end{subfigure}
\centering\caption{{\bf Display order affects image memorability.} 
This figure shows the target images from two different orders of the same set of images (target and fillers), along with their memorability scores (Mem) and the standard deviation (SD).
Depending on the context, the memorability score can dramatically change, e.g., the kitchen.
Conversely, the lighthouse, which is unique in both sequences, gets the same memorability score.
}
\label{fig:displayOrder}
\end{figure*}

We measured the correlation between different groups of observers.
When shown sets of different order, the correlation was low $\rho=0.4$, $\sigma=0.2$, 
while when shown sets of the the same order, the correlation was high $\rho=0.7$, $\sigma=0.1$.
An example of why this happens is displayed in Figure~\ref{fig:displayOrder}.
Take for instance, the kitchen scene. 
When presented after different scenes (top), its memorability score is $0.66$, however, when presented after a sequence of bedrooms (bottom), its memorability score jumped to $0.98$.

This suggests that the order of images in the sequence should not be overlooked.
The current practice of taking random orders and averaging might alleviates the influence of this extrinsic effect, but it does not consider it fully.
In fact, it averages sequences where the image is memorable with sequences where it is not, making the mean scores an inaccurate representation.
We conclude that a single score per image is not a sufficient representation.
\vspace{-0.1in}
\section{Conclusion}
This paper has studied the relation between convolutional neural networks for image classification and memorability prediction. 
It introduced {\em MemBoost}, a network that reaches the limit of human performance on the largest existing dataset for image memorability, {\em LaMem}.

{\em MemBoost} is based on three key observations:
(i) As object classification CNNs improve, so does image memorability prediction.
(ii) Scene classification plays a bigger role in memorability prediction than object classification, but their combination is preferable.
(iii) It suffices to train a regression layer on top of a CNN for object \& scene recognition to achieve on par results with those attained by re-training the entire CNN.
These observations were examined one-by-one via extensive experiments and were thoroughly analyzed.

Since our network already achieves human performance on LaMem, the next stage in memorability prediction is to produce a larger, more challenging dataset.
When doing so, various factors should be re-considered.
We provide some guidelines for designing the next-generation memorability dataset.
These guidelines regard the number of observers, the validity of maintaining one score per image, and the need  to re-think the order of images.
In the future, more factors may be studied.

\appendix
\section{Dataset details}
Tables~\ref{tbl:datasets}-\ref{tbl:train-datasets} provide the details on the 
datasets used throughout the paper.
\vspace{-0.1in}
\begin{table}[htb]
\begin{center}
\small{
\begin{tabular}{| l | c | c | c |}
\hline
 & \multicolumn{3}{ c |}{\textbf{Memorability Datasets}}  \\ 
\cline{2-4}
\textbf{Properties} & \textbf{LaMem \cite{khosla2015understanding}} & \textbf{SUN-Mem \cite{isola2011makes}} & \textbf{Figrim \cite{bylinskii2015intrinsic}} \\
\hline
Data type 
& Objects \& scenes & 397 Scenes & 21 scenes \\ \hline
\# Images 
& 58,741  &  2,222 & 1,754\\ \hline
Mean memorability 
& 75.6\(\pm\)12.4  &  67.5\(\pm\)13.6 & 66\(\pm\)13.9 \\ \hline
Human consistency 
& 0.68  & 0.75  & 0.74 \\ \hline
\end{tabular}
}
\end{center}
\centering\caption{{\bf Memorability datasets.} 
The  memorability scores are the mean and standard deviation over the entire dataset.
The consistency values are the average of the Spearman Rank Correlation between different groups of observers.}
\vspace*{-0.3cm}
\label{tbl:datasets}
\end{table}

\begin{table}[htb]
\begin{center}
\vspace{-0.3in}
\begin{tabular}{| l | c |}
\hline
\textbf{Training Datasets} & \textbf{Data type}\\ \hline
{ImageNet}~\cite{deng2009imagenet} & 1000 objects \\ \hline
{Places205}~\cite{zhou2014learning} & 205 scenes \\ \hline
{Places365}~\cite{zhou2017places} & 365 scenes \\ \hline
{Hybrid1205}~\cite{zhou2014learning}=ImageNet+Places205& 1000 objects + 205 scenes \\ \hline
{Hybrid1365}~\cite{zhou2017places}=ImageNet+Places365  & 1000 objects + 365 scenes\\ \hline
\end{tabular}
\end{center}
\centering\caption{{\bf Training datasets of images} of objects, of scenes and of both.
}
\vspace*{-0.4cm}
\label{tbl:train-datasets}
\end{table}
 
\newpage
{\small
\bibliographystyle{splncs}
\bibliography{bibliography}
}
\end{document}